\relax
\documentclass[letterpaper]{article} 
\usepackage{main}  
\usepackage{times}  
\usepackage{helvet} 
\usepackage{courier}  
\usepackage[hyphens]{url}  
\usepackage{times}
\usepackage{soul}
\usepackage{url}
\usepackage[utf8]{inputenc}
\usepackage[small]{caption}
\usepackage{amsmath}
\usepackage{booktabs}
\usepackage{algorithm}
\usepackage{algorithmic}
\usepackage{times}  
\usepackage{helvet}  
\usepackage{courier}  
\usepackage{url}  

\usepackage[algo2e,ruled,vlined,linesnumbered]{algorithm2e}

\usepackage{latexsym} 

\usepackage{booktabs} 
\usepackage{amssymb}
\usepackage{color}

\usepackage[algo2e,ruled,vlined,linesnumbered]{algorithm2e}
\SetKw{KwBy}{by}

\usepackage{graphicx} 
\usepackage{subfigure}
\usepackage{multirow}

\makeatletter
\newcommand{\printfnsymbol}[1]{%
  \textsuperscript{\@fnsymbol{#1}}%
}
\makeatother

\title{Customized Graph Embedding: \\Tailoring Embedding Vectors to different Applications}

\author{
\Large \textbf{Bitan Hou\textsuperscript{\rm 1\thanks{The work was partially done when the author visited MSRA.}\thanks{Equal Contribution}}, Yujing Wang\textsuperscript{\rm 2,3\printfnsymbol{2}}, Ming Zeng\textsuperscript{\rm 4\printfnsymbol{2}}} \\
\Large \textbf{Shan Jiang\textsuperscript{\rm 5}, Ole J. Mengshoel\textsuperscript{\rm 4}, Yunhai Tong\textsuperscript{\rm 3}, Jing Bai\textsuperscript{\rm 2}} \\
\textsuperscript{\rm 1}Shanghai Jiao Tong University $~~~$ 
\textsuperscript{\rm 2}Microsoft Research Asia \\
\textsuperscript{\rm 3}Key Laboratory of Machine Perception, MOE, School of EECS, Peking University \\
\textsuperscript{\rm 4}Carnegie Mellon University $~~~$
\textsuperscript{\rm 5}University of Illinois at Urbana-Champaign \\
houbitan@sjtu.edu.cn;$~~~$ \{yujwang, jbai\}@microsoft.com;$~~~$ yhtong@pku.edu.cn\\
\{ming.zeng, ole.mengshoel\}@sv.cmu.edu;$~~~$ sjiang18@illinois.edu; $~~~$ jbai@microsoft.com\\
}
\begin{document}

\maketitle

\begin{abstract}
Graph is a natural representation of data for a variety of real-word applications, such as knowledge graph mining, social network analysis and biological network comparison. For these applications, graph embedding is crucial as it provides vector representations of the graph. One limitation of existing graph embedding methods is that their embedding optimization procedures are disconnected from the target application. In this paper, we propose a novel approach, namely Customized Graph Embedding (CGE) to tackle this problem. The CGE algorithm learns customized vector representations of graph nodes by differentiating the importance of distinct graph paths automatically for a specific application. Extensive experiments were carried out on a diverse set of node classification datasets, which demonstrate strong performances of CGE and provide deep insights into the model.
\end{abstract}

\section{Introduction}
Graph is a natural format of data representation in many real-world scenarios \cite{goyal2018survey}, including knowledge graph mining~\cite{shi2017proje}, social network analysis~\cite{zhang2017user} and biological network comparison~\cite{zhu2013increasing}. Graph embedding methods generate dense vector representations of the graph and benefit a variety of downstream applications like node classification, link prediction and graph categorization. Existing methods of graph embedding can be classified into three categories, namely factorization-based~\cite{goyal2018survey}, random walk-based~\cite{perozzi2014deepwalk} and deep learning-based~\cite{kipf2016semi} algorithms. In this paper, we focus on random walk-based solutions because they are efficient and effective for large-scale graphs in real-world applications ~\cite{perozzi2014deepwalk}. 

The path sampling procedure is crucial for random walk-based graph embedding algorithms. Intuitively, different applications require different path sampling strategies to emphasize specific information in a graph. Suppose that we want to enhance the performance of medical Q\&A by exploiting Wikipedia knowledge graph. Wikipedia contains general knowledge which is not related to the medical domain. Such knowledge can be de-emphasized in the embedding training procedure to generate tailored node embedding vectors. However, none of the state-of-the-art methods provide a systematical solution to optimize path sampling and embedding training procedures jointly~\cite{cai2018survey}.

In this paper, we propose Customized Graph Embedding (CGE) to tackle the aforementioned problem. CGE samples a set of paths randomly from the graph and re-weights them through a neural network model before embedding training. The customized embedding procedure can be formulated as a bi-level optimization problem~\cite{sinha2018review} which consists of two loops. In the inner loop, node embedding vectors and supervised model parameters are trained based on a fixed re-weighting strategy; while in the outer loop, the goal is to find a re-weighting strategy to minimize the semi-supervised loss. We propose two neural network architectures as the re-weighting model, \emph{i.e.}, Convolutional Neural Network~\cite{krizhevsky2012imagenet} and Long Short-Term Memory~\cite{graves2005framewise}. We also implement a simple re-weighting strategy based on average pooling for comparison. Following~\cite{yang2016revisiting}, we develop both transductive and inductive variants of the CGE approach, where the transductive algorithm only deals with available nodes in the training phrase, and the inductive one can be generalized to unseen nodes. The major contributions of this paper are summarized as below. 

\begin{itemize}
    \item To the best of our knowledge, GCE is the first attempt to generate tailored node embedding vectors by capturing task-oriented information in a graph. 
    \item CGE achieves state-of-the-art performances on four node classification datasets (CITESEER, CORA, PUBMED and NELL) that are widely used for the empirical evaluation of graph embedding methods.
    \item We provide deep insights into CGE to explain the benefits of path re-weighting. We observe that CGE automatically retrieves task-oriented paths from the graph. Moreover, the re-weighting scores are highly related to the length and node diversity of each path.
\end{itemize}

\section{Related Work}
Graph embedding benefits a wide range of graph-related applications. The advantage of graph embedding is to incorporate all kinds of information from the graph into dense vectors, which can be leveraged by downstream applications as advanced features. For example, \cite{zhang2017learning} proposes a deep feature learning paradigm by mining the visual-semantic embeddings from noisy, sparse, and diverse social image collections. It demonstrates superior performance in various applications such as content-based retrieval, classification, and image captioning. 

As summarized in ~\cite{cai2018survey}, there are mainly four kinds of input graphs for embedding applications: homogeneous graph, heterogeneous graph, graph with auxiliary information, and graph constructed from non-relational data. Homogeneous graph (e.g., Webpage link graph) is a basic form where every node or edge corresponds to the same type. Heterogeneous graph is another common setting where nodes and edges are from multiple categories. For instance, \cite{zhao2015representation} constructs a knowledge graph from Wikipedia which consists of three types of nodes (i.e., entity, category, and word) and three types of edges (i.e., entity-entity, entity-category, and word-word). One can also build graphs with auxiliary information for each node. For example, \cite{dai2016discriminative} leverages attributes, labels, and text descriptions to improve the quality of embedding vectors. In addition, the graph can be constructed from non-relational data. As an example, \cite{yan2007graph} constructs an intrinsic graph and a penalty graph to capture intra-class compactness and inter-class separability respectively.

From the perspective of embedding output, existing algorithms can be categorized in to four kinds: node embedding, edge embedding, hybrid embedding and whole-graph embedding~\cite{cai2018survey}. Node embedding generates a representation vector for each node; edge embedding aims at learning a dense vector for each edge, hybrid embedding jointly embeds a combination of different types of graph components~\cite{cavallari2017learning}; and whole-graph embedding typically learns the dense representation for a small graph (e.g., proteins and molecules)~\cite{nikolentzos2017matching}. This paper mainly focuses on node embedding, while the idea of customized graph embedding can be easily generalized to other kinds of algorithms.

Moreover, the algorithms of graph embedding can be categorized into three broad groups, \emph{i.e.}, factorization-based, random walk-based and deep learning-based models~\cite{goyal2018survey}. Factorization-based algorithms represent the connection between nodes in a matrix and calculate embedding vectors through matrix factorization. The limitation of factorization-based methods is that they are prohibitive on large graphs, because the procedures of proximity matrix construction and eigen-decomposition are time- and space-consuming~\cite{demmel2007fast}. Random walk-based methods are more efficient and scalable for large graphs and have been extensively studied in both academia and industry. As a pioneering work, DeepWalk~\cite{perozzi2014deepwalk} samples a set of paths randomly from the graph and leverages SkipGram~\cite{mikolov2013distributed} to calculate node representation vectors. However, it ignores the label information in embedding learning procedure so that the representation vectors can not be customized to a specific application. In Node2vec~\cite{grover2016node2vec}, the sampling strategy can be adaptive to BFS or DFS style based on different applications, but it is not expressive enough for fine-grained sampling. Planetoid~\cite{yang2016revisiting} is a semi-supervised graph embedding framework that jointly optimizes the node label and graph context. Unfortunately, its path sampling strategy is still not task-specific. In addition, there are deep learning-based approaches that apply auto-encoders~\cite{wang2016structural} or Convolutional Neural Networks~\cite{niepert2016learning} on the graph directly. Graph Convolutional Network (GCN)~\cite{kipf2016semi} is a representative approach of this category, which achieves superior results on node classification tasks. 

\section{Customized Graph Embedding}

\begin{figure}[ht]
\centering
\includegraphics[height=4.0in]{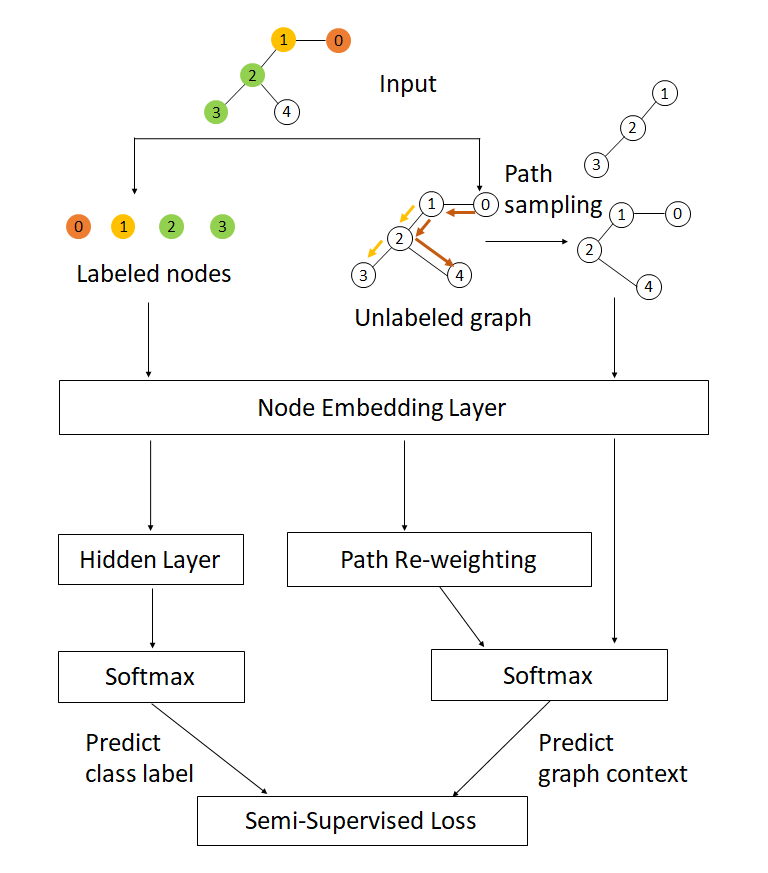}
\caption{The overall architecture of CGE.}
\label{fig:whole}
\end{figure}

The overall architecture of Customized Graph Embedding (CGE) is illustrated in Figure~\ref{fig:whole}. The input is a large graph with labeled nodes of the downstream application. Similar to traditional random walk-based approaches, we sample a collection of sub-paths from the input graph. The most salient part proposed in CGE is to re-weight each sub-path by a neural network model to reflect its importance to a specific task. Then, the unsupervised loss is calculated by the weighted paths and the supervised loss is formulated by labeled nodes. Finally, we perform gradient decent on the semi-supervised loss function to optimize node embedding vectors. 

\subsection{Path Sampling \& Pair Generation}

The sampled paths are generated by a random walk on the entire graph. We take each node as a starting point and sample the next node uniformly from its neighbours. This sampling process is repeated multiple times until the path is long enough. Then, we extract sub-paths from each path within a fixed sliding window. For example, if the original path is $n_1n_2n_3n_4n_5$, and the sliding window size is 3, all the sub-path extracted should be $n_1$; $n_2$; $n_3$; $n_4$; $n_5$; $n_1n_2$; $n_2n_3$; $n_3n_4$; $n_4n_5$; $n_1n_2n_3$; $n_2n_3n_4$; and $n_3n_4n_5$. The starting and ending node in each sub-path formulates a node pair. Following \cite{yang2016revisiting}, we also sample a collection of pairs consisting of nodes with the same training label. We do not re-weight these pairs because they are all highly relevant to the final target. 

\subsection{Path Re-weighting}

In unsupervised graph embedding methods such as DeepWalk~\cite{perozzi2014deepwalk}, paths are usually assumed to be equally weighted in the loss function. However, the embedding vectors learned in this way are not always optimal for an arbitrary application. Take Wikipedia knowledge graph as an example. Each path in the graph reflects a piece of knowledge in a particular domain; and a random path may be noisy or irrelevant to the target application. Thus, we re-weight the graph paths to indicate their importance scores so that the essential knowledge for a specific task can be emphasized.

Our re-weighting model is based on a neural network architecture. We denote the re-weighting model as $A$, which takes a sequence of nodes in the path as input and outputs an importance weight in the range of $[0, 1)$. The framework is quite general and multiple architectures are applicable for the re-weighting model. Here we employ two neural network architectures, \emph{i.e.}, Convolutional Neural Network (CNN) and Long Short-Term Memory (LSTM). For comparison, we also implement a baseline re-weighting strategy based on average pooling. The three re-weighting models are described as follows:

\begin{itemize}
\item \textbf{Average Pooling}: It calculates the average embedding vectors of all nodes in the path. A feed-forward layer is applied before sigmoid activation. 
\item \textbf{CNN}: 
Two 1-D convolution layers are stacked on the input vector as shown in Equation (\ref{equation:cnn}). In each convolutional layer, the filter size is set to be 3 and the number of filters is 1. The weight of path $\mathbf{p}_k$ is given by:

\begin{align}
  &  \mathbf{h}_k^1 = \text{Conv1D}(\mathbf{p}_k, \mathbf{w}^1) \nonumber \\
  &  h_k^2 = \text{Conv1D}(\mathbf{h}_k^1, \mathbf{w}^2) \nonumber \\
  &  A(\mathbf{p}_k) = \text{sigmoid}(h_k^2)
\label{equation:cnn}
\end{align}
where $\mathbf{w}^1$ and $\mathbf{w}^2$ are the parameters in the first and second layers respectively. $\mathbf{h}_k^1$ is a vector, whose dimension equals to the embedding size, and $h_k^2$ is a scalar.

\item \textbf{LSTM}: LSTM is a natural choice for modeling sequential input~\cite{hochreiter1997long}. In our implementation, the hidden state of LSTM is passed to a feed-forward layer and then to a sigmoid output layer. The weight of path $\mathbf{p}_k$ can be calculated by:
\begin{align}
&\mathbf{h}_k = \text{LSTM}(\mathbf{p}_k, \mathbf{w}) \nonumber \\
&v_k = (\mathbf{w}_{\text{linear}})^T \mathbf{h}_k \nonumber \\
&A(\mathbf{p}_k) = \text{sigmoid}(v_k) \label{eq:A}
\end{align}

where $\mathbf{w}$ and $\mathbf{h}_k$ are the parameters and hidden states of LSTM model respectively; 

$\mathbf{w}_{\text{linear}}$ is the vector of parameters of the feed-forward layer. 

\end{itemize}

\subsection{Loss Function}
The loss function is semi-supervised and consists of two parts, \emph{i.e.}, the supervised loss $L_s$ and unsupervised loss $L_u$. 

The supervised loss optimizes the prediction of the target variables explicitly, while the unsupervised loss is used as a regularization. The semi-supervised loss is formulated as:
\begin{align}
&L(A, \mathbf{e}, \mathbf{\theta}) = L_s + \lambda L_u,
\end{align}
where $\lambda$ is a hyper-parameter that controls the trade-off between $L_s$ and $L_u$; $\mathbf{\theta}$ denotes the set of trainable parameters in the supervised model; and $\mathbf{e}$ is the final embedding output.

\subsubsection{Unsupervised Loss}
Given an input graph, our goal is to learn the embedding representation of each node in the graph tailored to a particular task. Learning the embedding vectors of nodes in a graph forms an unsupervised graph embedding problem which can be solved by DeepWalk~\cite{perozzi2014deepwalk}, where the loss function is similar to Skip-Gram~\cite{mikolov2013distributed}. We generalize the loss function of previous work by enabling importance weighting on each path:

\begin{align}
L_u = \sum_{k}A(\mathbf{p}_k) \text{loss}(\mathbf{w}_i, \mathbf{e}_j)
\end{align}
where $\mathbf{p}_k$ represents a sub-path sampled randomly from the graph; $A(\mathbf{p}_k)$ calculates the importance weight of $\mathbf{p}_k$; $i$ and $j$ denote the start node and end node of $\mathbf{p}_k$ respectively; $\mathbf{w}_i$ represents the input embedding vector of node $i$; $\mathbf{e}_j$ is the output vector of node $j$; and $\text{loss}(\mathbf{w}_i, \mathbf{e}_j)$ is the loss of predicting node $j$ by node $i$, which is formulated as:

\begin{align} \label{eq:negative_loss}
& \text{loss}(\mathbf{w}_i, \mathbf{e}_j) \nonumber = -\log{p(\mathbf{e}_j|\mathbf{w}_i)} \\ 
& ~~~~~~~~~~~~~~ = -\log{\frac{\exp({\mathbf{e}_j^{\intercal}\mathbf{w}_i})}{\sum_{k=1}^{K}{\exp({\mathbf{e}_k^{\intercal}\mathbf{w}_i})}}}
\end{align}
where $K$ is the number of all distinct nodes. In practice, if the graph size is too large, we can apply negative sampling~\cite{mikolov2013distributed} to accelerate the training process. 

\subsubsection{Supervised Loss}
Based on node embedding vectors $\mathbf{e}_i$, the supervised loss of node classification is formulated as below:
\begin{align}
\label{eq:transductive}
L_s = \sum_{i}l(y_i, f(\mathbf{x}_i, \mathbf{e}_i, \mathbf{\theta})),
\end{align}
where $\mathbf{x}_i$ and $\mathbf{e}_i$ represent for the input feature vector and output embedding vector of node $i$ respectively; $f$ denotes the supervised model, and $\mathbf{\theta}$ is the parameters of the model. The ground truth label for node $i$ is denoted by $y_i$, and $l(\cdot, \cdot)$ is the loss function. In our experiments, we utilize cross-entropy loss in node classification tasks. 

We implement two flavors of the supervised loss, i.e., transductive and inductive~\cite{yang2016revisiting}. In the transductive setting, the embedding vector $\mathbf{e}_i$ is trained on the training data and retrieved directly when the same node appears in the test phase. The drawback is that it cannot be generalized to unseen nodes in the training phase.
To alleviate this problem, we also propose an inductive algorithm, which learns an embedding model to project any input feature vector $\mathbf{x}_i$ to an output embedding vector $\mathbf{e}(\mathbf{x}_i)$. 

\begin{itemize}
    \item \textbf{Transductive Setting}: the embedding vectors $\textbf{e}_i$ serves as additional features to $\mathbf{x}_i$ and can be learnt jointly in the semi-supervised model. The transductive semi-supervised loss can be written as:
    \begin{align}
    \label{eq:supervise}
    & L_s = \sum_{i}l(y_i, f(\mathbf{x}_i, \mathbf{e}_i, \mathbf{\theta})), \\
    & f(\mathbf{x}_i, \mathbf{e}_i, \mathbf{\theta}) = \text{softmax}(f(h^k(\mathbf{x}_i) \oplus h^l(\mathbf{e}_i))) \label{eq:supervised_f}
    \end{align}
    where $h^k$ and $h^l$ are single feed-forward layers, taking the feature vectors and embedding vectors as input respectively; $\theta$ represents the learnable parameters of layer $h^k$ and $h^l$. The results of feed-forward layers are concatenated before calculating the prediction result by linear projection $f$ and softmax activation.
    \item \textbf{Inductive Setting}: the embedding vector $\mathbf{e}_i$ is calculated as a function of input feature vector $\mathbf{x}_i$ in order to be generalized to unseen instances. The transductive semi-supervised loss can be written as:
    \begin{align}
    \label{eq:inductive}
    & L_s = \sum_{i}l(y_i, f(\mathbf{x}_i, \mathbf{e}(\mathbf{x}_i), \mathbf{\theta})), \\
    & f(\mathbf{x}_i, \mathbf{e}(\mathbf{x}_i), \mathbf{\theta}) = \text{softmax}(h(\mathbf{x}_i \oplus \mathbf{e}(\mathbf{x}_i))
    \end{align}
    where $h$ is a single feed-forward layer, taking the concatenation of feature vector $\mathbf{x}_i$ and embedding vector $\mathbf{e}(\mathbf{x}_i)$ as input; $\mathbf{\theta}$ stands for the learnable parameters in layer $h$. In our experiments, the embedding function $\mathbf{e}(\mathbf{x}_i)$ is implemented as a single feed-forward layer. 
\end{itemize}

\subsection{Optimization}
The goal of customized graph embedding is to automatically find the optimal function $A^*$ that minimizes the validation loss $L(A^*,\mathbf{e}^*, \mathbf{\theta}^*)$ where the embedding vector $\mathbf{e}^*$ and model parameter $\mathbf{\theta}^*$ are learned jointly to fit the training data. It can be naturally formulated as a bi-level optimization problem: 
\begin{align}
\label{eq:bi-level}
 \min_A \qquad & L_{\text{val}}(A, \mathbf{e}^*, \mathbf{\theta}^*) \nonumber \\
 s.t. \qquad & \mathbf{e}^*, \mathbf{\theta}^*   =  \operatorname*{argmin}_{\mathbf{e}, \mathbf{\theta}} L_{\text{train}}(\mathbf{w}_A, \mathbf{e}, \mathbf{\theta})
\end{align}
where $A$ stands for the re-weighting strategy; $\mathbf{e}$ and $\theta$ stands for the embedding and model parameters separately. Let $\mathbf{\alpha} = (\mathbf{e}, \mathbf{\theta})$ and $\mathbf{w}_A$ denotes the parameters in the re-weighting model $A$, Equation (\ref{eq:bi-level}) can be rewritten as:
\begin{align}
\label{eq:optimization}
\min_{\mathbf{w}_A} \qquad & L_{\text{val}}(\mathbf{w}_A, \mathbf{\alpha}^*) \nonumber \\
s.t. \qquad & \alpha^* = \operatorname*{argmin}_{\alpha} L_{\text{train}}(\mathbf{w}_A, \alpha)
\end{align}

Solving the bi-level optimization problem in (\ref{eq:optimization}) exactly is prohibitive, because of the nested structure: the optimal value of $\alpha^*$ needs to be recomputed whenever $\mathbf{w}_A$ has any change. We thus utilize an approximate iterative optimization procedure similar to~\cite{darts2018}, where $\alpha$ and $\mathbf{w}_A$ are updated alternately. First, we update $\alpha=(\mathbf{e}, \mathbf{\theta})$ for a single step towards minimizing the training loss $L_{\text{train}}(\mathbf{w}_A, \alpha)$. Then, keeping the embedding vector $\mathbf{e}$ and supervised model parameter $\mathbf{\theta}$ fixed, we update the parameter $\mathbf{w}_A$ (which controls the weighting strategy of sampled paths) towards minimizing the validation loss:
\begin{align}
\label{eq:valid_loss}
L_{\text{val}}(\mathbf{w}_A, \alpha - \xi\nabla_{\alpha}L_{\text{train}}(\mathbf{w}_A,\alpha))
\end{align}
where $\xi$ is the learning rate of a virtual gradient step. The idea behind virtual gradient step is to find a weighting strategy $\mathbf{w}_A$ which has low validation loss when the supervised parameters $\alpha^*(\mathbf{w}_A)$ are optimized. Here the one-step unroll weights serves as a surrogate for the optimal value of $\alpha^*(\mathbf{w}_A)$. 

In the optimization procedure, the embedding and model parameters are updated alternatively according to the following gradients:
\begin{align}
\label{derivation_1}
& \nabla_{\alpha}L_{\text{train}}(\mathbf{w}_A,\alpha) = \nabla_{\alpha}L_{s,\text{train}}(\mathbf{w}_A,\alpha) + \nabla_{\alpha}L_{u}(\mathbf{w}_A,\alpha)
\end{align}
\begin{align}
\label{derivation_2}
& \nabla_{\mathbf{w}_A}L_{\text{val}}(\mathbf{w}_A, \alpha -\xi\nabla_{\alpha}L_{\text{train}}(\mathbf{w}_A,\alpha)) \nonumber \\ 
& ~~~~~~~~~ = \nabla_{\mathbf{w}_A}L_{u}(\mathbf{w}_A, \alpha -\xi\nabla_{\alpha}L_{\text{train}}(\mathbf{w}_A,\alpha))  
\end{align}
Here $L_{s, \text{train}}$ represents the supervised loss on the training nodes, $L_{u}$ is the unsupervised loss on the sampled pairs, which does not differentiate between $L_{\text{train}}$ and $L_{\text{val}}$.
Note that the supervised loss does not have derivation to $\mathbf{w}_A$, so the first item $L_{s, \text{val}}$ in Equation (\ref{derivation_2}) is ignored.

Taking the transductive setting for example, the overall procedure of CGE is summarized in Algorithm~\ref{alg:bo}. The inputs include labeled nodes, unlabeled nodes and graph edges, while the outputs are node embedding vectors.

\begin{algorithm2e}
\SetKwInOut{Input}{Input}\SetKwInOut{Output}{Output}
  \Input{graph $G$, labeled data $\mathbf{x}_{1:n_L}, y_{1:n_L}$, unlabeled data $\mathbf{x}_{n_L:n_L+n_U}$; \\
  parameters of weighting model $\mathbf{w}_A$; \\
  parameters of supervised model $\alpha=(\mathbf{e}, \mathbf{\theta})$; \\ 
  hyper-parameter $\lambda$, $\xi$}
  \Output{embedding vector $\mathbf{e}$; model parameters $\mathbf{w}_A$, $\mathbf{\theta}$}
  Initialize embedding vector and model parameters \\
  Sampling a collection of paths randomly. \\
  \While{not converged}{
  Update $\alpha$ by descending $\nabla_{\alpha}L_{s,\text{train}}(\mathbf{w}_A,\alpha) + \nabla_{\alpha}L_{u}(\mathbf{w}_A,\alpha)$ \\
  Update $\mathbf{w}_A$ by descending $\nabla_{\mathbf{w}_A}L_{u}(\mathbf{w}_A, \alpha -\xi\nabla_{\alpha}L_{\text{train}}(\mathbf{w}_A,\alpha)$
  }
 \caption{Transductive algorithm of CGE}\label{alg:bo}
\end{algorithm2e}

\section{Experiments}
\subsection{Datasets}
We have six datasets derived from four corpus of node classificaiton, the statistics of which are shown in Table~\ref{table:stat}). In CITECEER, CORA and PUBMED\footnote{Datasets are available from: \url{https://linqs.soe.ucsc.edu/data}}, we leverage 20 instances in each class for training and 1000 instances in each class for validation and test. The NELL (Never Ending Language Learning) corpus is built on NELL knowledge base ~\cite{carlson2010toward} and a hierarchical entity classification dataset ~\cite{dalvi2016hierarchical}. 
The goal is to classify the entities into one of 210 possible categories. Following \cite{yang2016revisiting}, we generate three datasets (NELL01, NELL001, NELL0001) from NELL corpus which contain 10\%, 1\% and 0.1\% labeled instances correspondingly\footnote{The datasets can be found at \url{http://www.cs.cmu.edu/~zhiliny/data/nell_data.tar.gz}}.
The official datasets do not provide a golden split of test and validation. In our experiments, we randomly select one half for test and another half for validation. The same split result is used for all experiments in this paper. 

\begin{table}[t]
\centering
\scalebox{0.8}{
\begin{tabular}{lcccccc}
\toprule
\centering
Corpus  & \#Classes & \#Nodes & \#Edges & \#Labeled & \#Valid\&Test \\
\midrule
CITESEER & 6         & 3,327   & 4,732 & 120 & 1,000  \\
CORA     & 7         & 2,708   & 5,429  & 140 & 1,000 \\
PUBMED   & 3         & 19,717  & 44,338 & 60 & 1,000  \\
NELL01	 & 210 		& 65,755 	 & 266,14 & 1,054 & 848	 \\
NELL001	 & 210 		& 65,755 	 & 266,14 & 161   & 987	 \\
NELL0001	 & 210 		& 65,755 	 & 266,14   & 105   & 969	 \\
\bottomrule
\end{tabular}
}
\caption{Six datasets used in our experiments.}
\label{table:stat}
\end{table}

\begin{table*}[t]
\centering
\scalebox{1.0}{
\begin{tabular}{c|l|cccccc}
\toprule
\multicolumn{1}{l|}{}        & Method      & \multicolumn{1}{l}{CITESEER} & \multicolumn{1}{l}{CORA} & \multicolumn{1}{l}{PUBMED} & \multicolumn{1}{l}{NELL01} & \multicolumn{1}{l}{NELL001} & \multicolumn{1}{l}{NELL0001} \\
\midrule
\multirow{2}{*}{Unsupervised} 
& DeepWalk & 0.610 & 0.667 & 0.749 & 0.619 & 0.426 & 0.205\\
& node2vec & 0.619 & 0.692 & 0.74 & 0.642 & 0.458 & 0.206\\
\midrule
\multirow{4}{*}{Transductive} 
& PLANETOID-T & 0.610& 0.718 & 0.737 & 0.724 & 0.488 & 0.266 \\
& CGE-AVERAGE & 0.622 & 0.745 & 0.691 & 0.629 & 0.417 & 0.208\\
& CGE-CNN     & \textbf{0.635}$^{\ast}$ & \textbf{0.761}$^{\ast}$ & 0.682 & 0.623 & 0.416 & 0.204 \\
& CGE-LSTM    & 0.629 & 0.740& \textbf{0.754}$^{\ast}$ & \textbf{0.726}& \textbf{0.514}$^{\ast}$ & \textbf{0.356}$^{\ast\ast}$ \\
\midrule
\multirow{4}{*}{Inductive}    
& PLANETOID-I & 0.679 & 0.674 &\textbf{0.804} & 0.631 & 0.433 & 0.200 \\
& CGE-AVERAGE & 0.667 & 0.672 & 0.792 & 0.614 & 0.416 & 0.191 \\
& CGE-CNN & 0.674 & 0.684 & 0.779 & 0.609 & 0.414 & 0.180 \\
& CGE-LSTM & \textbf{0.691}& \textbf{0.692}$^{\ast}$ & 0.800 & \textbf{0.664}$^{\ast}$ & \textbf{0.475}$^{\ast}$ & \textbf{0.295}$^{\ast}$\\
\bottomrule
\end{tabular}
}
\caption{Accuracy of the unsupervised baselines (DeepWalk, node2vec), state-of-the-art semi-supervised method (PLATOID-T/I) and Customized Graph Embedding (CGE) models (CGE-Average, CGE-CNN, CGE-LSTM). PLANETOID-T and PLANETOID-I denote transductive and inductive semi-supervised graph embedding respectively. Significance tests show that our CGE-based results are significantly better than the results of PLANETOID-T and -I respectively with $p < 0.05$ (marked by a star~$^\ast$) and $p < 0.01$ (marked by two stars~$^{\ast\ast}$) .
}
\label{table:main_result}
\end{table*}

\subsection{Experimental Settings}
For all datasets, we adopt accuracy as the evaluation metric. We compare CGE with Planetoid~\cite{yang2016revisiting} in the transductive and inductive settings respectively\footnote{We search for the best hyper-parameters of both algorithms on the validation set.}, which is the state-of-the-art solution for semi-supervised graph embedding. We also adopt two commonly-used unsupervised approaches, DeepWalk~\cite{perozzi2014deepwalk} and node2vec~\cite{grover2016node2vec} as baseline solutions. Three variants of CGE are evaluated in the experiments, namely \textbf{CGE-Average} (reweighting with average pooling), \textbf{CGE-CNN} (reweighting with CNN), and \textbf{CGE-LSTM} (reweighting with LSTM). In our experiments, the embedding sizes of CITESEER, CORA, PUBMED and NELL datasets are set as 50, 135, 128 and 256 respectively. In the path sampling procedure, the max path length is set to 10. Sliding window size is 3 by default, but we change it to 10 when analysing the correlation between path weights and path lengths. SGD is used for bi-level optimization. 

\subsection{Comparison Results}
The comparison results are shown in Table~\ref{table:main_result}. On all the six datasets, the transductive CGE methods outperform previous SOTAs consistently. In particular, comparing to PLANETOID-T, CGE-CNN achieves $2.5\%$ and $4.3\%$ improvements in accuracy on CITESEER and CORA datasets; CGE-LSTM achieves $1.7\%$ improvement on PUBMED, $0.2\%$, $2.6\%$ and $9\%$ lifts on NELL01, NELL001 and NELL0001 datasets respectively. In the inductive setting, CGE-LSTM achieves on-par performance with PLANETOID-I on PUBMED and outperforms the best state-of-the-art results by a large margin on the other five datasets. 
These results suggest that the CGE-based approaches effectively learn better embedding representations that significantly improve the performance of node classification. In addition, LSTM turns to be the best re-weighting model.

\subsection{Combination with GCN}

\begin{table}[!ht]
\centering
\begin{tabular}{lccc}
\toprule
                  & CITESEER & CORA  & PUBMED \\
\midrule
GCN Baseline      & 0.720    & 0.813 & 0.792  \\
GCN + CGE    & \textbf{0.747}$^{\ast}$    & \textbf{0.827}$^{\ast}$ & \textbf{0.846}$^{\ast}$  \\
\bottomrule
\end{tabular}
\caption{Accuracy of GCN baseline and CGN + CGE. Significance tests show that GCN+CGE results are significantly better than the results of GCN baseline with $p < 0.05$ (marked by a star~$^\ast$).}
\label{table:GCN}
\end{table}

Graph Convolutional Network (GCN) is a new state-of-the-art method for semi-supervised node classification on graphs~\cite{kipf2016semi}. It encodes the graph structure using a neural network model and trains the supervised target on all labeled nodes. As this method does not calculate node embedding vectors, it is not directly comparable to other graph embedding approaches. Instead, we combine GCN with CGE for node classification. Specifically, the softmax output from CGE is used as an additional feature for each node in the GCN model. As shown in Table~\ref{table:GCN}, the output of CGE provides extra information to the original GCN model, establishing new state-of-art performances on multiple datasets.

\begin{figure}[hbt!]
\centering
\includegraphics[height=1.9in]{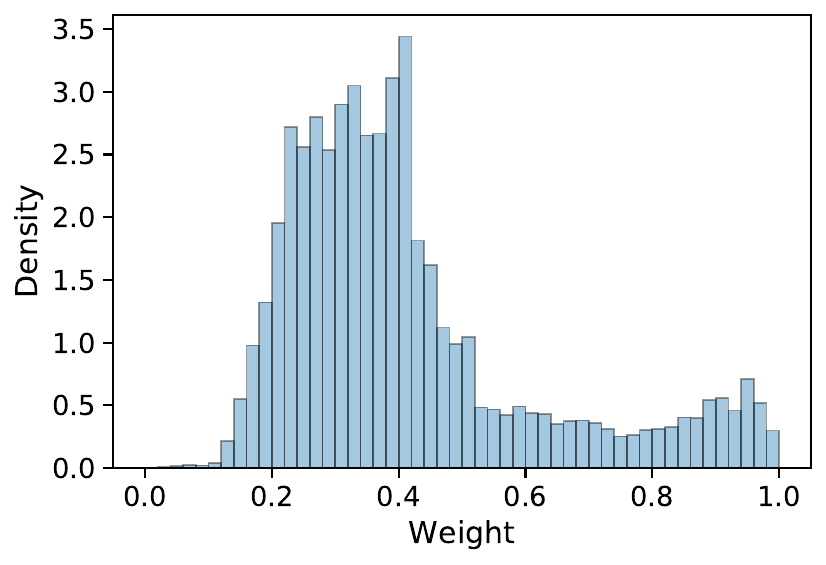}
\caption{Comparison of average weighting scores of paths with / without masked labels.}
\label{fig:distribution}
\end{figure}

\begin{figure}[hbt!]
\centering
\includegraphics[height=2.0in]{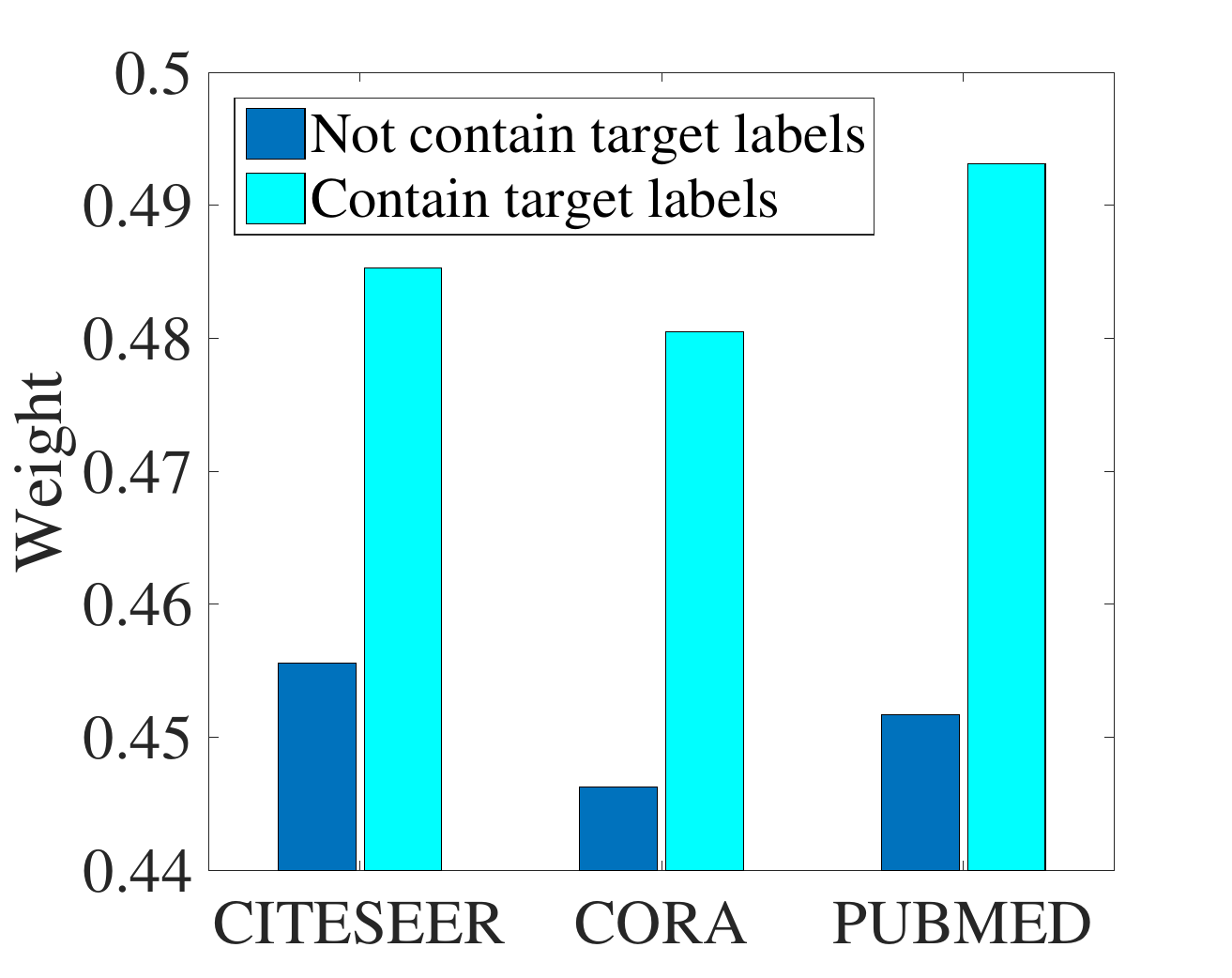}
\caption{Comparison of average weighting scores of paths with / without masked labels.}
\label{fig:weight_relevant}
\end{figure}

\subsection{Analysis}
The re-weighting scores learned by GCE have considerable variance to distinguish their importance to the downstream task. For example, the distribution of re-weighting scores of CGE-LSTM on the CITESEER dataset is visualized in Figure \ref{fig:distribution}. 
To understand how this benefits the performance, we provide more analysis on the results of CGE-LSTM. 

\subsubsection{Path Relevance}
To examine if CGE selects the most relevant paths automatically, we randomly mask a half of target node labels in the training phase. Then, we train a CGE model and analyze the re-weighting scores of sampled paths. We categorize the paths into two groups, i.e., \textit{containing the target label} or \textit{without the target label}. Obviously, paths containing the target labels should be more relevant than the paths without target labels. As shown in Figure~\ref{fig:weight_relevant}, the results agree well with our intuition that more relevant paths can be emphasized by the CGE algorithm automatically. 

\begin{table}

	\begin{minipage}{\linewidth}
		\centering
		\includegraphics[height=2.0in]{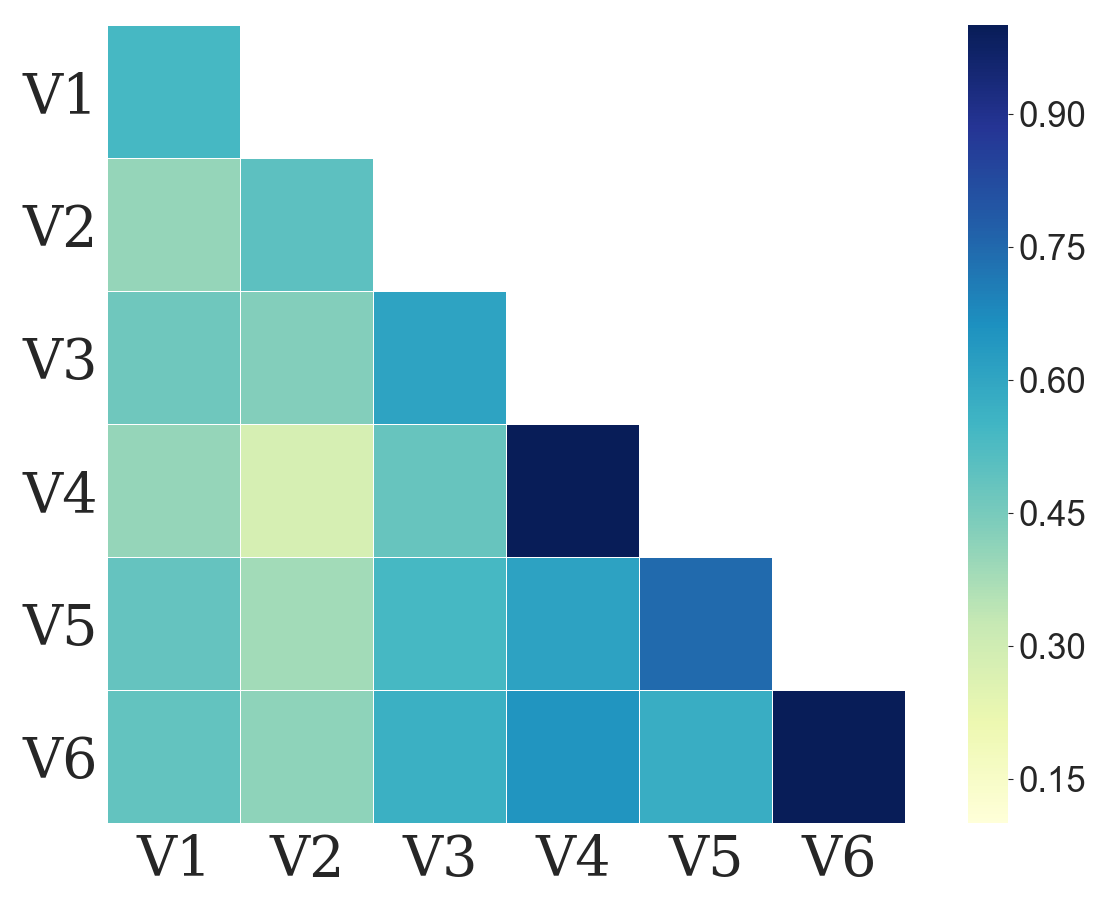}
		\captionof{figure}{Visualization of relevance between different categories.}
		\label{fig:heatmap}
	\end{minipage}\vfill 
	\vspace{0.5cm}

	\begin{minipage}{\linewidth}
		\centering
        \begin{tabular}{ccc}
        \toprule
        Label & Venue & Number\\
        \midrule
        V1 & Communications of ACM & 817\\
        V2 & Computer & 254\\
        V3 & IEEE Transactions on Computer & 776\\
        V4 & Discrete Mathematics & 10\\
        V5 & Theoretical Computer Science & 114\\
        V6 & Computational and Applied Mathematics & 36\\
        \bottomrule
        \end{tabular}
        \caption{Summary of paper citation graph}
        \label{table:AE}
	\end{minipage}
	
\end{table}

\begin{figure*}[hbt!]
\centering
\includegraphics[height=1.6in]{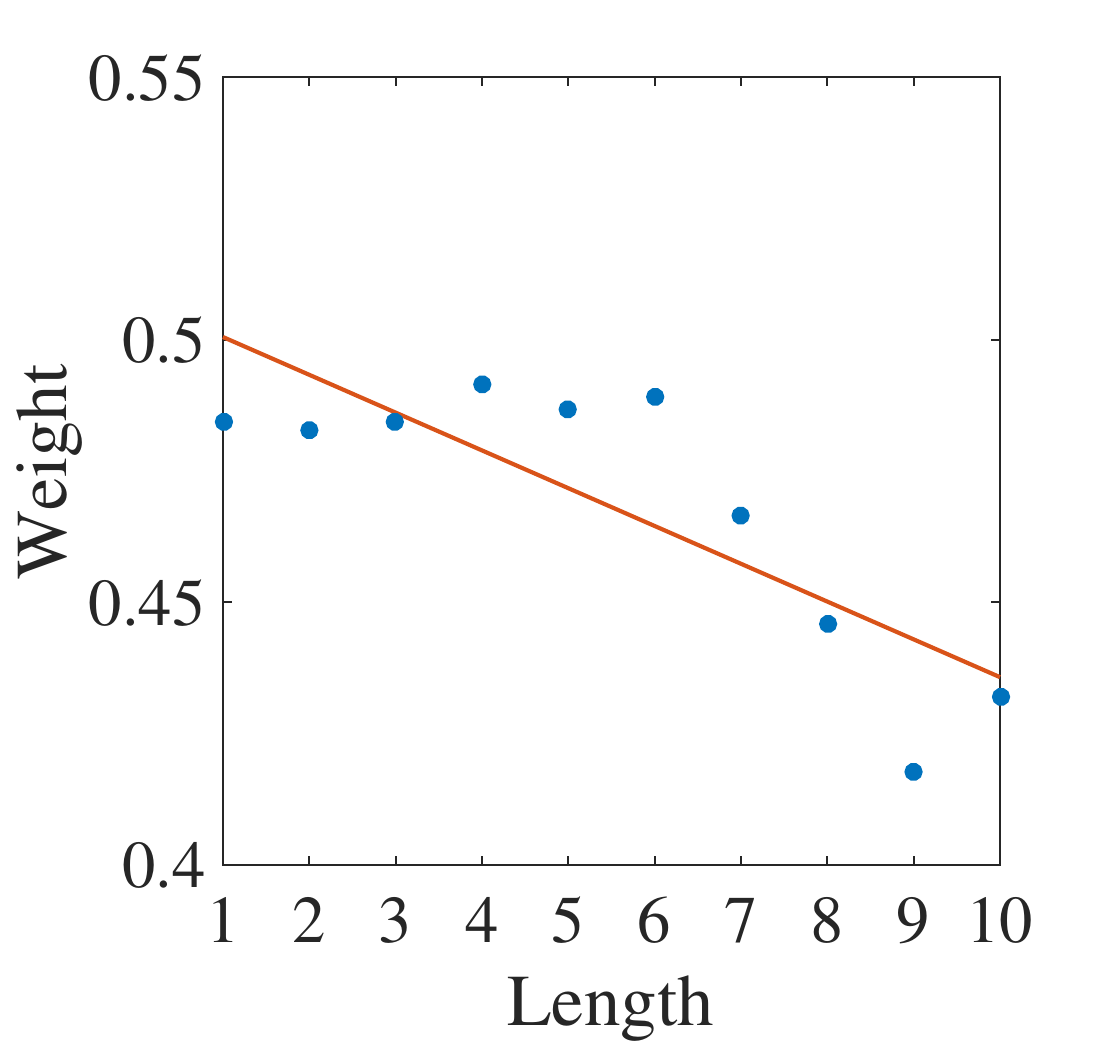}
\includegraphics[height=1.6in]{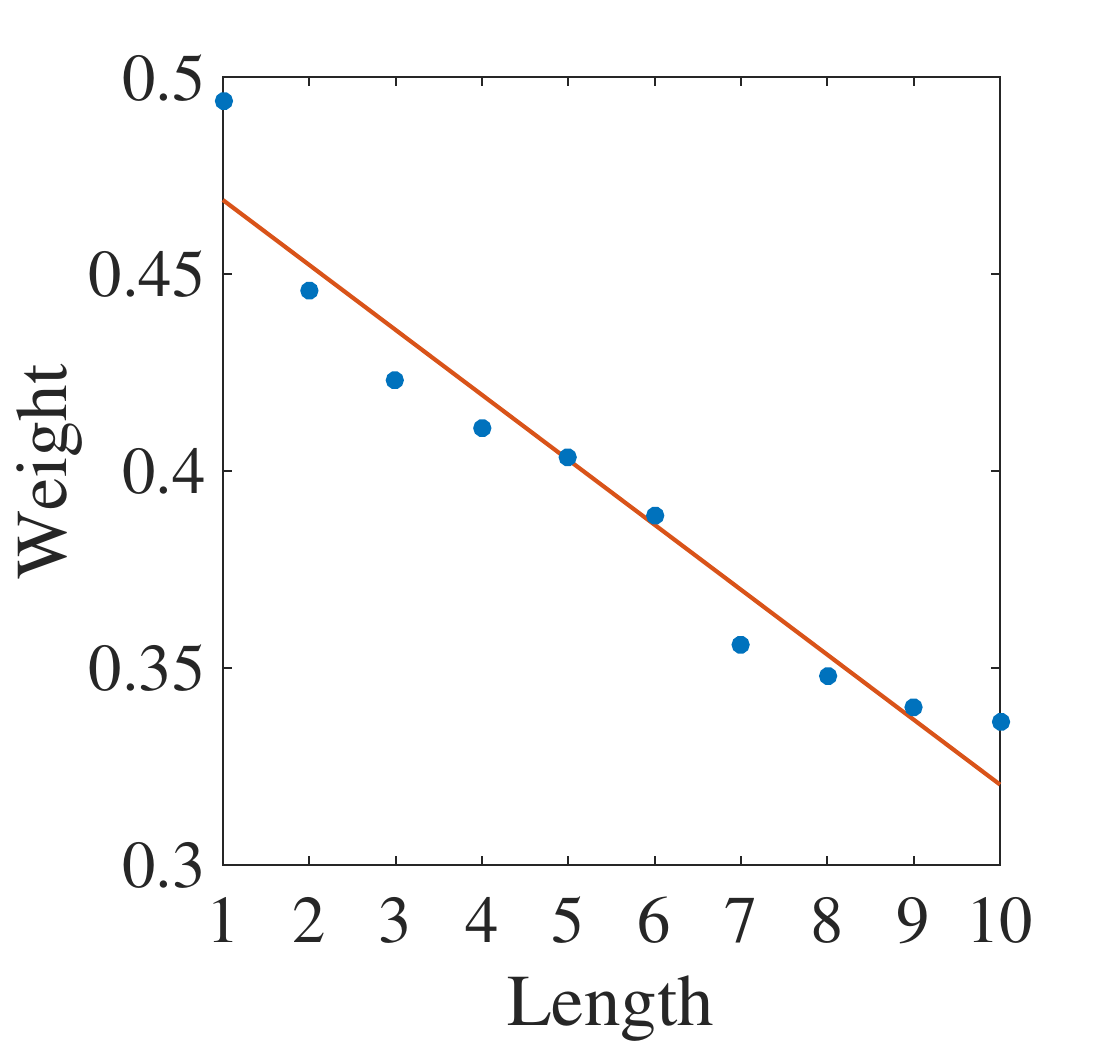}
\includegraphics[height=1.6in]{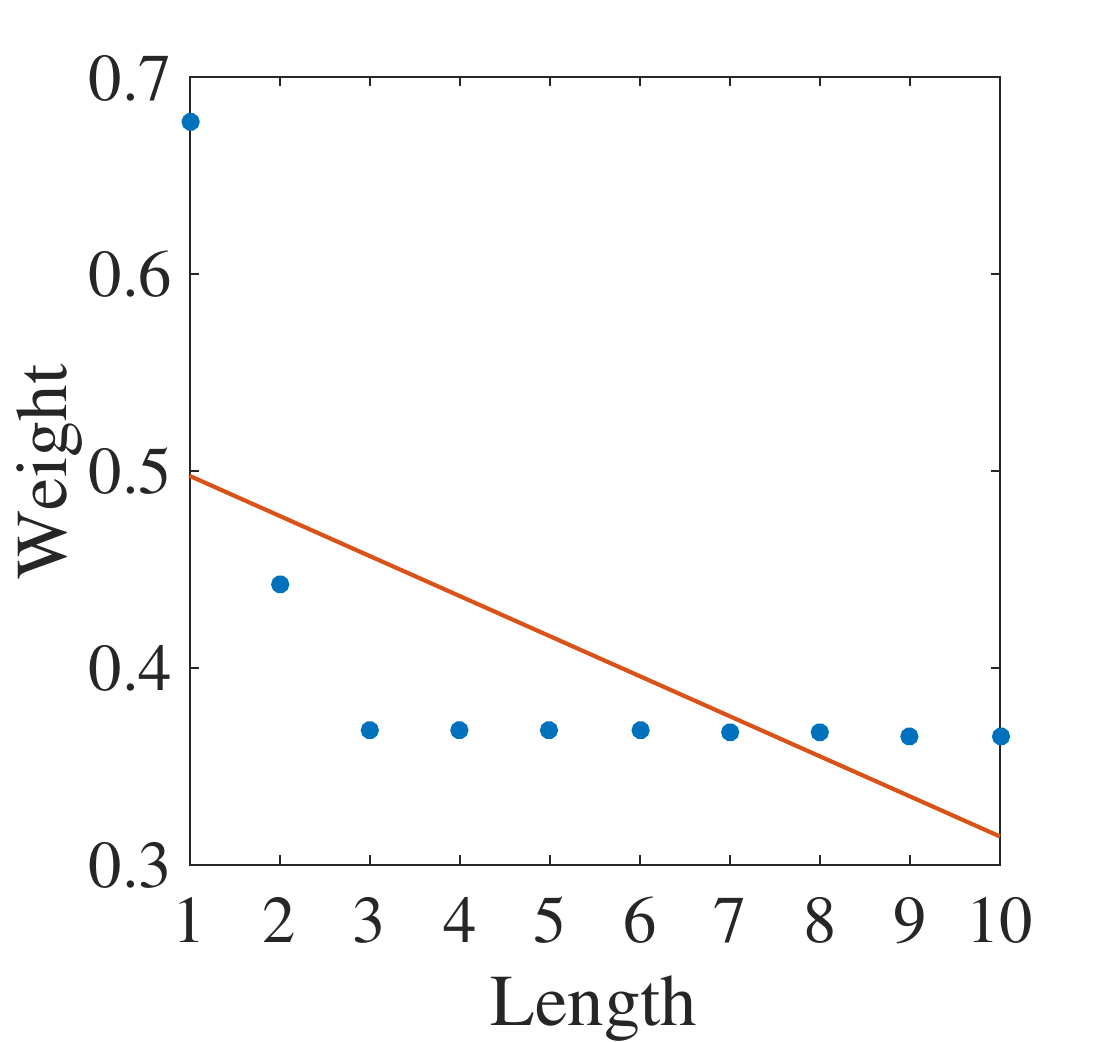}
\caption{Correlation between path weight and node distance from left to right: (a) CITESEER ($r=-0.82$) (b) CORA ($r=-0.97$) (c) PUBMED ($r=-0.62$).}
\label{fig:weight_distance}
\end{figure*}
\begin{figure*}[hbt!]
\centering
\includegraphics[height=1.6in]{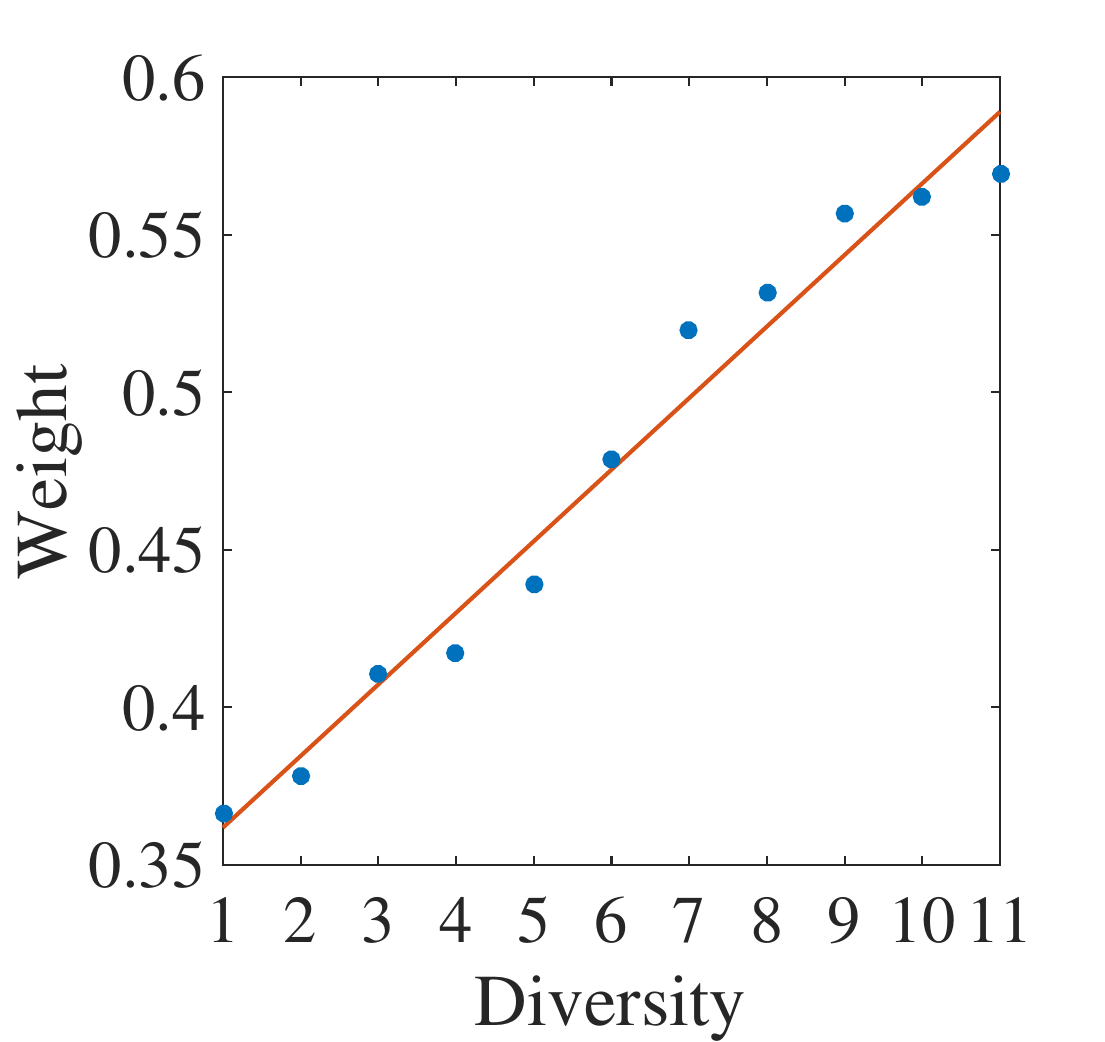}
\includegraphics[height=1.6in]{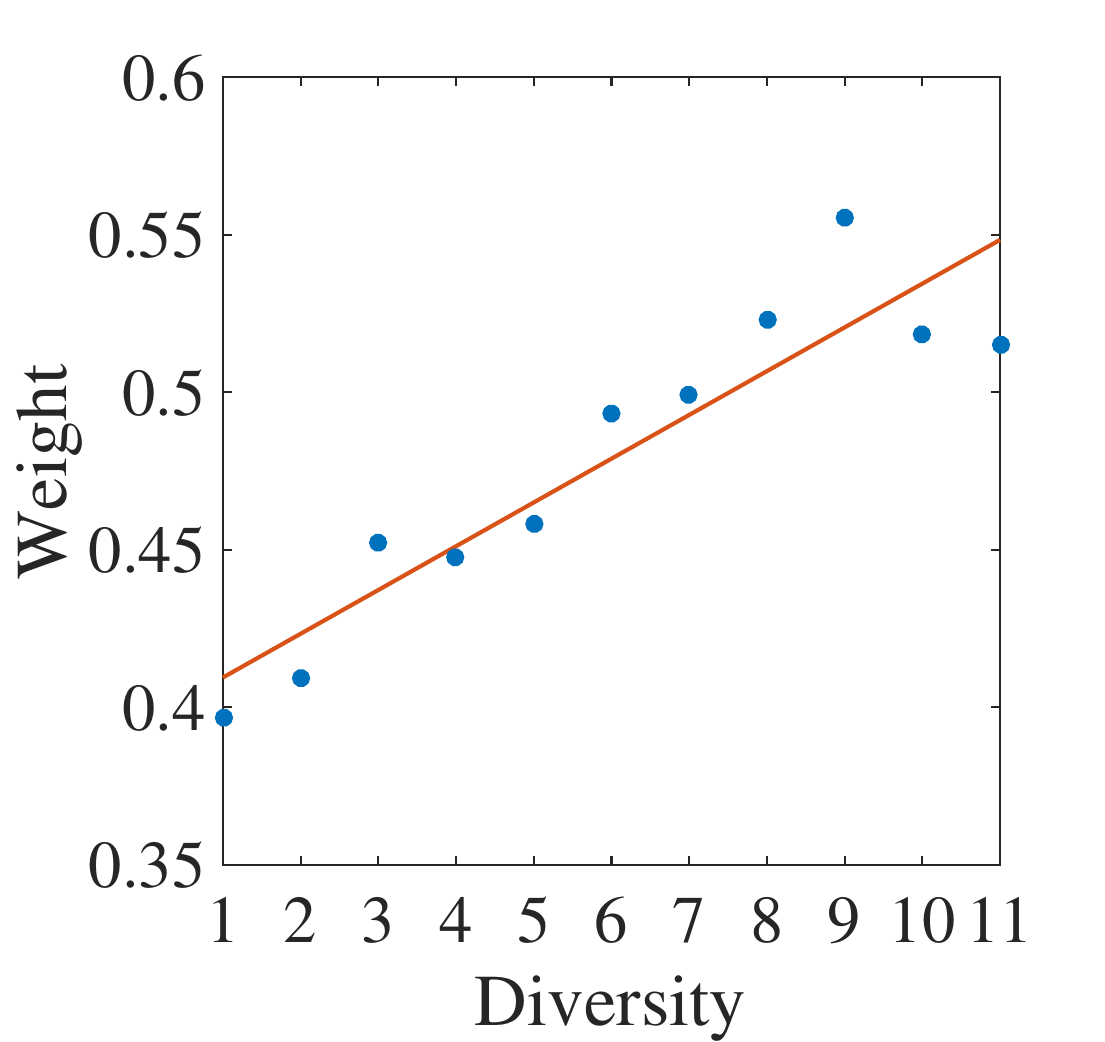}
\includegraphics[height=1.6in]{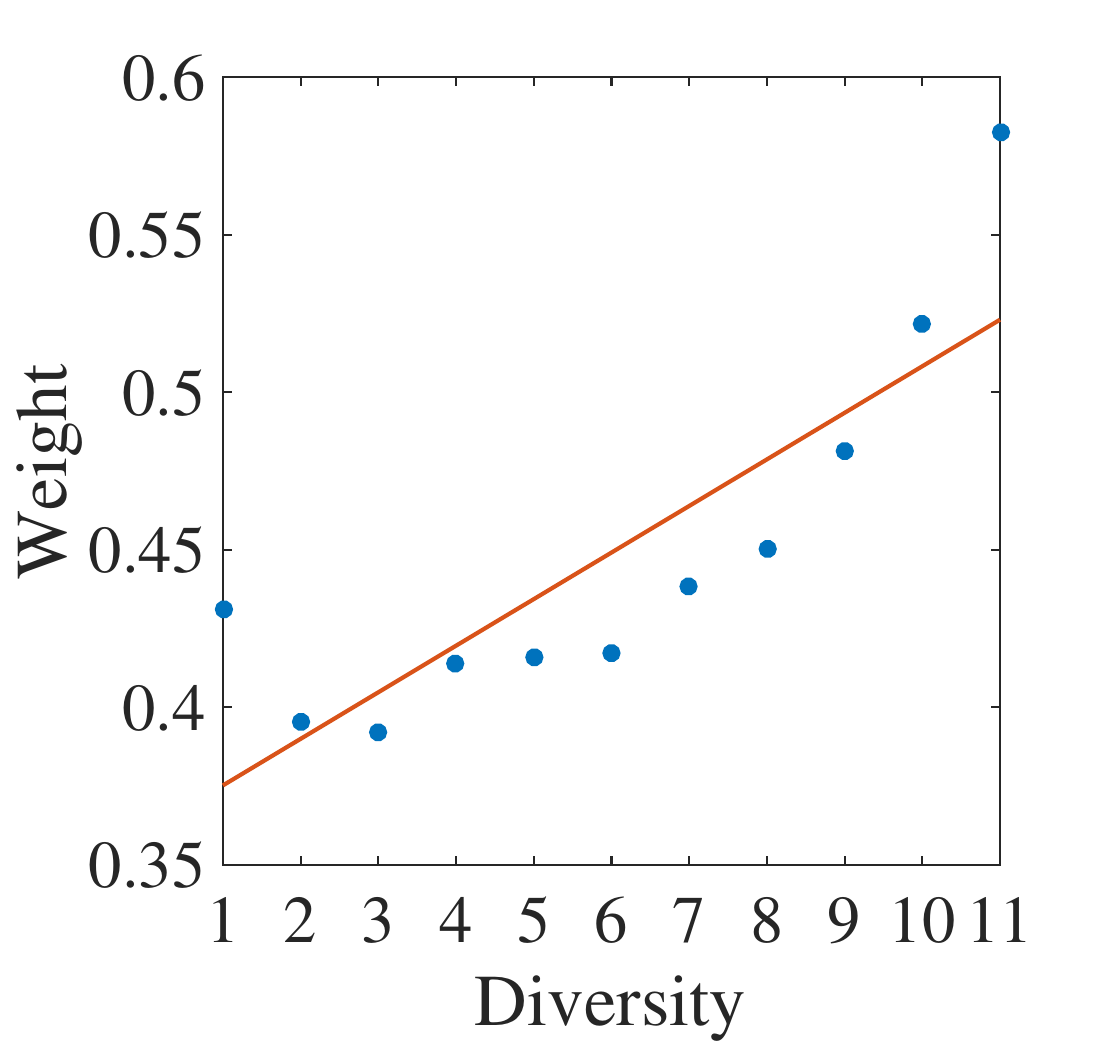}
\caption{Correlation between path weight and path diversity from left to right: (a) CITESEER ($r=0.98$) (b) CORA ($r=0.92$) (c) PUBMED ($r=0.84$).}
\label{fig:weight_diversity}
\end{figure*}

\subsubsection{Semantics Explanation}

In order to analysis the semantic correlations between different categories, we build another sub-graph from a paper citation network\footnote{The datasets can be found at \url{https://www.aminer.cn/citation}}~\cite{Tang2008citation}. Each graph node belongs to one of 6 classes (venues) as summarized in Table~\ref{table:AE}. Based on this dataset, we perform 6 binary classification tasks where one venue is treated as positive and others as treated as negative. At the same time, each path can be classified into respective groups by considering if it contains at least one node with a certain venue. Taking \textit{Discrete Mathematics} as an example, the target task is to identify if an arbitrary node belongs to this category or not. After training the model, we calculate the average weighting scores of paths containing each venue (e.g, \textit{Computational and Applied Mathematics}). This can be seen as a relevance indicator of two corresponding venues, i.e., \textit{Discrete Mathematics} and \textit{Computational and Applied Mathematics} in our example.

Figure~\ref{fig:heatmap} visualizes the average weighting scores of paths associated with category $i$ when taking category $j$ as the classification target. This figure shows an analogy of pairwise attentions, where a larger score indicates higher relevance between corresponding classes. As shown in the figure, the inter-class relevance learned by the re-weighting model is interpretable. For example, the venues of \textit{Discrete Mathematics} and \textit{Computational and Applied Mathematics} are highly correlated with each other. Another finding is that each node has the strongest correlation with itself. Moreover, the average weights of \textit{Communications of ACM}, \textit{Computer} and \textit{IEEE Transactions on Computer} are relatively low. It is reasonable because their papers are from broader areas. To conclude, CGE has the capability of underlining relationships between different categories even though it does not known the exact semantic meaning. 
\subsubsection{Weight Correlation}

We visualize the correlations between re-weighting scores and two major properties of paths, \emph{i.e.}, length and diversity. For a path $(n_i, n_{i+1}, ..., n_j)$, its length is $j-i+1$ and diversity is defined as the number of distinct nodes in the path. The correlations between path weight and length are visualized in Figure~\ref{fig:weight_distance}, where $x$-axis denotes path length and $y$-axis denotes the average weighting score of the corresponding paths. As illustrated by the figure, the re-weighting strategy learns that the correlation between nodes decays as the distance between them becomes larger. Figure~\ref{fig:weight_diversity} shows the impact of node diversity on path weights. We observe that path weight increases with the enlargement of node diversity. It seems that the paths containing more distinct nodes are more informative for model training, thus they are emphasized by the CGE model.

\section{Conclusion}
In this paper, we introduce a novel framework for semi-supervised graph embedding, namely Customized Graph Embedding (CGE). In contrast to previous semi-supervised graph embedding approaches, our method is capable of differentiating the importance of different graph paths for learning node embedding vectors. The experimental results demonstrate significant performance enhancement over state-of-the-art methods. Further analysis shows that the CGE algorithm automatically underlines the most relevance paths in a graph. In addition, we analyze the correlations between weighting scores and specific properties of graph paths. The correlations learned by the model agree well with human common-sense. One direction of future work is to extend the proposed framework to support edge and sub-graph embedding. We also plan to apply this framework to applications in industry, such as search relevance and query recommendation.

\bibliographystyle{main}
\bibliography{main}

\end{document}